\documentclass[oneside,english]{article}
\usepackage[LGR,T1]{fontenc}
\usepackage[utf8]{inputenc}
\usepackage[a4paper]{geometry}
\usepackage{color}
\usepackage[ngerman]{babel}
\usepackage{array}
\usepackage{varioref}
\usepackage{float}
\usepackage{units}
\usepackage{textcomp}
\usepackage{amsthm}
\usepackage{amsmath}
\usepackage{amssymb}
\usepackage{amsfonts}
\usepackage{graphicx}
\usepackage{setspace}
\usepackage{nomencl}
\usepackage{umlaute}
\usepackage{enumitem}
\usepackage{datetime}

\usepackage[notocbib]{apacite}

\usepackage[bottom,hang]{footmisc}
\usepackage{geometry}
\usepackage{ifthen}
\usepackage{nomencl}

\makenomenclature
\renewcommand{\nomgroup}[1]{%
\renewcommand{\makelabel}[1]{##1}
\item[~]

\ifthenelse{\equal{#1}{D}}{%
\item[\textbf{Distributions}]}{%

\ifthenelse{\equal{#1}{S}}{%
\item[\textbf{Sets and spaces}]}{%

\ifthenelse{\equal{#1}{F}}{%
\item[\textbf{Functions}]}{%

\ifthenelse{\equal{#1}{M}}{%
\item[\textbf{Miscellaneous}]}{%

\ifthenelse{\equal{#1}{O}}{%
\item[\textbf{Observations}]}{%

\ifthenelse{\equal{#1}{L}}{%
\item[\textbf{Losses and risks}]}{%

\ifthenelse{\equal{#1}{C}}{%
\item[\textbf{Cardinalities}]}{%

}}}}}}}%
\item[~]
\let\makelabel\nomlabel
}

\newcommand{\R}{\mathbb{R}}

\newcommand{\N}{\mathbb{N}}

\newcommand{\IF}{\textup{IF}}
\newcommand{\epslim}{{\lim_{\varepsilon \searrow 0}}}

\newcommand{\BorelXY}{{\mathfrak{B}_{\X\times\Y}}}

\newcommand{\BorelXbY}{{\mathfrak{B}_{\Xb\times\Y}}}

\newcommand{\Pb}{{{P}_{b}}}
\newcommand{\Pbtildeepsilonz}{{{{\tilde P}_{b,\varepsilon,z}}}}
\newcommand{\PbtildeepsilonQb}{{{{\tilde P}_{b,\varepsilon,Q}}}}
\newcommand{\PX}{{{P}^\mathcal{X}}}
\newcommand{\PbX}{{ {P}_b^{\mathcal{X}_b} }}

\newcommand{\Y}{{\mathcal{Y}}}
\newcommand{\X}{{\mathcal{X}}}
\newcommand{\Xb}{{\mathcal{X}_{b}}}

\newcommand{\frakDn}{{{D}_n}}                       
\newcommand{\frakDnb}{{{D}_{n,b}}}              

\newcommand{\calDn}{{\mathcal{D}_n}}                        
\newcommand{\calDnb}{{\mathcal{D}_{n,b}}}             

\newcommand{\bsumme}{{\sum\limits_{b=1}^{B}}}

\newcommand{\lambdab}{{\lambda_b}}
\newcommand{\lambdanb}{{\lambda_{(n_b,b)}}}

\newcommand{\fLPlambdacomp}{{f_{\Ls,P,\lambda}^{comp}}} 
\newcommand{\fLDnlambdacomp}{{f_{\Ls,\frakDn,\lambda}^{comp}}} 
 
\newcommand{\fLQlambdacomp}{{f_{\Ls,Q,\lambda}^{comp}}}

\newcommand{\fLDnlambdancomp}{{f_{\Ls,\frakDn,\lambda_n}^{comp}}}

\newcommand{\Ls}{{ {L^*}   }}

\newcommand{\lokaltheoretisch}{{f_{b,\Ls,\Pb,\lambdab}}} 
\newcommand{\lokalempirisch}{{f_{b,\Ls,\frakDnb,\lambdab}}} 
\newcommand{\zusammentheoretischx}{{ {\bsumme w_b(x) \lokaltheoretisch(x) }}}

\newcommand{\zusammenempirischx}{{ {\bsumme w_b(x) \lokalempirisch(x) }}}

\newcommand{\nto}{{\ \xrightarrow[{\tiny n\to\infty}]{}\ }}

\newcommand{\btoB}{{\left\{1,\ldots,B\right\}}}

\newtheoremstyle{break}
	  {10pt}
 	  {10pt}
 	  {\itshape}
 	  {\parindent}
 	  {\bfseries}
 	  {}
 	  {\newline}
 	  {}
\theoremstyle{break}

\addto\captionsngerman{
\renewcommand{\bibname}{References}
\renewcommand{\refname}{References}

}

\newtheorem{stz}{Satz}[section]
\newtheorem{dfi}[stz]{Definition}

\newtheorem{bsp}[stz]{Example}
\newtheorem{cor}[stz]{Corollary}

\newtheorem{thm}[stz]{Theorem} 
\newtheorem{prp}[stz]{Proposition}

\newenvironment{Proof}{\begin{proof}[Proof.]}{\end{proof}}

\usepackage{remreset}
\makeatletter
\@removefromreset{footnote}{section}
\makeatother

\title{\textbf{Quantitative Robustness of Localized Support Vector Machines}}
 
\author{\textbf{Florian Dumpert}\\
\normalsize{Department of Mathematics, University of Bayreuth, Germany}}
\date{}

\begin{document} 

\selectlanguage{english}

\maketitle
 
\begin{abstract}\noindent
The huge amount of available data nowadays is a challenge for kernel-based machine learning algorithms like SVMs with respect to runtime and storage capacities. Local approaches might help to relieve these issues and to improve statistical accuracy. It has already been shown that these local approaches are consistent and robust in a basic sense. This article refines the analysis of robustness properties towards the so-called influence function which expresses the differentiability of the learning method: We show that there is a differentiable dependency of our locally learned predictor on the underlying distribution. The assumptions of the proven theorems can be verified without knowing anything about this distribution. This makes the results interesting also from an applied point of view.

\end{abstract}  
  



\section{Introduction}


This paper analyzes a special robustness property of localized kernel-based, non-parametric statistical machine learning methods, in particular of support vector machines (SVMs)  \fullcite{boser1992training, cortes1995support}, and methods close to them. There are many general introductions to these methods from the view of computer science and statistics. Summarizing textbooks are for example \fullciteA{cristianini2000introduction}, \fullciteA{scholkopf2001learning}, \fullciteA{cucker2007learning}, or \fullciteA{steinwart2008support}. These methods became pretty popular in many fields of science, see for example \fullciteA{ma2014support}. The analysis provided by this paper refers to supervised learning, i.\,e. to classification or regression problems. Beyond this, support vector machines are a suitable method for unsupervised learning (e.\,g. novelty detection), too.


The paper can be seen as a sequel to \fullciteA{dumpert2018neuro} where universal consistency and robustness with respect to the maxbias of localized support vector machines have already been shown. This paper is dedicated to refine the robustness analysis. It is organized as follows: Section~\ref{lmusvm} gives a short overview on support vector machines, Section~\ref{local} introduces shortly the idea of local approaches. The results concerning the influence function of localized support vector machines are given in Section~\ref{results}. Section~\ref{summary} finally summarizes the paper.

\section{Prerequisites}

\subsection{Support Vector Machines}\label{lmusvm}

A support vector machine is a minimizer of one of the following expressions,
\begin{align}
\mathcal R_{\X,\Ls,D_n,\lambda_n}(f) := \frac 1 n \sum\limits_{i=1}^n \Ls(y_i,f(x_i)) + \lambda_n \|f\|_H^2,  \label{emprisk}
\end{align}
\begin{align}
\mathcal R_{\X,\Ls,P,\lambda_n}(f) :=  \int\limits_{\X\times\Y} \Ls(y_i,f(x_i))\ dP(x,y) + \lambda_n \|f\|_H^2, \label{theorisk}
\end{align}
where (\ref{emprisk}) is called the empirical risk of a function $f$ with respect to a shifted loss function $\Ls$ and an empirical measure $D_n = n^{-1} \sum_{i=1}^{n} \delta_{(x_i,y_i)}$ (where $\delta_{(x,y)}$ denotes the Dirac measure at a point $(x,y)\in\X\times\Y$) based on a sample $\mathcal D_n =  \left(  (x_1, y_1), \ldots, (x_n, y_n) \right)$ of i.\,i.\,d. realizations (with respect to a joint distribution $P$ on $(\X\times\Y,\BorelXY)$, where $\BorelXY$ denotes the Borel-$\sigma$-algebra on $\X\times\Y$) of random variables $X$ (input, with values in $\X$) and $Y$ (output, with values in $\Y$). (\ref{theorisk}) is called the theoretical risk associated with  (\ref{emprisk}). Minimizers of (\ref{emprisk}) are called empirical support vector machines and will be denoted by $f_{\Ls,D_n,\lambda_n}$; minimizers of  (\ref{theorisk}), i.\,e. theoretical SVMs, will be denoted by $f_{\Ls,P,\lambda_n}$. A supervised loss function (or shorter: a loss function) has to measure the difference between observed and predicted values in an appropriate way and is defined as a measurable function $L:\Y\times\R \to [0,\infty[$. (For unsupervised learning a slightly different definition is needed.) In order to create the link to \fullciteA{dumpert2018neuro} we are also interested in the so-called shifted version $\Ls$ of a loss function $L$, defined by $\Ls:\Y\times \R \to \R, \ \Ls(y,t) := L(y,t) - L(y,0)$, see also Appendix B of \fullciteA{dumpert2018neuro}. Within the next lines, we have to recall some definitions and results. A loss function $L$ is called  convex, if $t\mapsto L(y,t)$ is  convex for all $y\in\Y.$ Its shifted version $\Ls$ is called  convex, if $t\mapsto \Ls(y,t)$ is  convex for all $y\in\Y.$ $L$ is called Lipschitz continuous if there is a constant $|L|_1\in [0,\infty[$ such that for all $y\in\Y$ and all $t, s\in \R$, $|L(y,t) - L(y,s)|\le|L|_1|t-s|$. Analogously, $\Ls$ is called Lipschitz continuous, if there is a constant $|\Ls|_1\in [0,\infty[$ such that for all $y\in\Y$ and all $t, s\in \R$, $|\Ls(y,t) - \Ls(y,s)|\le|\Ls|_1|t-s|$. It is easy to show that if $L$ is a loss function which is  convex, then $\Ls$ is  convex --- and if $L$ is a loss function which is Lipschitz continuous, then $\Ls$ is Lipschitz continuous with the same Lipschitz constant. Note that in all situations where the theoretical SVM with respect to an unshifted loss function $L$ ($f_{L,P,\lambda_n}$) exists, it holds true that $f_{L,P,\lambda_n} = f_{\Ls,P,\lambda_n}$. It is always true that $f_{L,D_n,\lambda_n} = f_{\Ls,D_n,\lambda_n}$. Hence, the (computational) algorithms and the resulting predictors are the same (as far as they exist) with or without shifting the loss function.

The regularization parameter $\lambda$ usually depends on the sample size $n\in\N$ (in this case, we write $\lambda_n$), is positive for all $n\in\N$, and plays an important role within the next sections.
The aim of support vector machines in supervised learning is to discover the influence of a (generally multivariate) input (or explanatory) variable $X$ on a univariate output (or response) variable $Y$. Our goal is to explore the functional relationship that describes the conditional distribution of $Y$ given $X$.
$\X$, the input space, is generally assumed to be a separable metric space. For some results of this paper $\X$ has additionally to be complete. \textbf{For the rest of the paper, the output space $\Y$ is assumed to be a closed subset of the real line }$\R$.
When we talk about a data set, a sample or observed data, we think (for $n\in\N$) about an $n$-tuple, but note that, although it is a tuple, we treat it like a set and use notations like $\in, \cap, ...$; nevertheless we allow that the sample contains a data point twice or several times. 
$H$ denotes a reproducing kernel Hilbert space (RKHS). For the  bijection between kernels and their reproducing kernel Hilbert spaces (RKHS) see \fullciteA{aronszajn1950}, \fullciteA{scholkopf2001learning} and \fullciteA{berlinet2001}. A very important connection between the functions in an RKHS and its corresponding kernel is given by the following propositions \fullcite[Lemma~4.23, Lemma~4.28]{steinwart2008support}.
\begin{prp}\label{dieungleichung}
A kernel $k$ is called  bounded if $\|k\|_\infty := \sup_{\substack{x\in\X}} \ \sqrt{k(x,x)} \ < \infty.$ If and only if the reproducing kernel $k$ of an RKHS $H$ is bounded, every $f\in H$ is bounded and for all $f\in H, x\in \X$ there is the inequality $|f(x)| = |\langle f, k(\cdot, x)\rangle_H| \le \|f\|_H \|k\|_\infty.$ Particularly:  $\|f\|_\infty\le \|f\|_H \|k\|_\infty.$
\end{prp}

\begin{prp}\label{stetigerkern}
Let $k$ be a kernel with RKHS $H$. Then $k$ is bounded and $k(\cdot,x):\X\to\R$ is continuous for all $x\in\X$ if and only if every $f\in H$ is bounded and continuous. Obviously: If $k(\cdot,\cdot)$ is continuous,  then $k(\cdot,x):\X\to\R$ is continuous for all $x\in\X$.
\end{prp}

SVMs are known to be universal (risk-)consistent, i.\,e.
$$\mathcal R_{\X,\Ls,P} (f_{\Ls,D_n,\lambda_n}) \nto \textup{inf}\left\{\mathcal R_{\X,\Ls,P}(f)\ |\ f:\X\to\R \text{ measurable}\right\} \ \text{in probability w.r.t.}\ P$$
under week assumptions ($\mathcal R_{\X,\Ls,P}(\cdot) := \mathcal R_{\X,\Ls,P,0}(\cdot)$).  For support vector machines concerning a shifted loss function we basically refer to \fullciteA{christmann2009consistency}.


\subsection{Localized approaches and regionalization}\label{local}

A short overview on the idea of localized statistical learning is already given in \fullciteA{dumpert2018neuro}. We now take it up again. There are two main aspects that show the need of localized approaches. First, the computational effort of kernel-based machine learning methods. The larger the sample the more costly the computation of a solution. Second, the statistical aspect.  Different areas of $\X\times\Y$ might have different claims on the statistical method: There might be regions that require simple functions serving as predictors while other regions might need more volatile functions. The success of  machine learning approaches often heavily depends on finding optimal hyperparameters. These parameters often determine the complexity of the predictor. By learning the set of hyperparameters for the whole input space, we often have to average out the specifics of the local areas. Local learning allows to use different hyperparameters and even different kernels in different regions. These regions have to fulfil some of the following assumptions.
\begin{itemize}
\item[\textbf{(R1)}]A regionalization method divides the input space $\X$ into possibly overlapping regions, i.\,e. $\X = \bigcup_{b=1}^{B_{n}}\X_{(n,b)}$ or $\X\times\Y = \bigcup_{b=1}^{B_{n}}\left(\X_{(n,b)}\times\Y\right)$. $B_{n}$ is the number of regions, usually chosen by the regionalization method and therefore depending (at least) on a subsample  drawn to do the regionalization. Note that $B :=B_{n}$ is constant after the regionalization, so we have $\X = \bigcup_{b=1}^{B}\X_{(n,b)}$ or $\X\times\Y = \bigcup_{b=1}^{B}\left(\X_{(n,b)}\times\Y\right)$. Note that this is not the same as \emph{robust learning from bites} \fullcite{bites2007}.

\item[\textbf{(R2)}] For every $b\in\btoB$ $\Xb$ is a separable metric space (which is easy to fulfil as subsets of separable sets are separable and subsets of metric spaces are metric spaces \fullcite[I.6.4, I.6.12]{DunfordSchwartz1958}), and, in addition, a complete measurable space, i.\,e., for all probability measures, $(\Xb\times\Y,\BorelXbY)$ is complete. Note that this  notion of completeness refers to the measurability of null sets, see \fullciteA[Definition~1.3.7]{ash2000probability}.

\item[\textbf{(R3)}] For $n\to\infty$, the regionalization method ensures $\left|\mathcal D_n \cap (\Xb\times\Y)\right |\to \infty$  for all $b\in\btoB$, i.\,e. $\lim_{\substack{n\to\infty}}\  \min_{b\in\btoB}\left|\mathcal D_n \cap (\Xb\times\Y) \right|= \infty.$ (For an arbitrary set $M$, $|M|$ denotes the number of its elements.)

\item[\textbf{(R4)}] Every region $\Xb$ is complete, $b\in\btoB$, in the sense that every Cauchy sequence in $\Xb$ has a limit in $\Xb$. Note that this is easy to ensure by using the completion of the results of the regionalization method. (This is not a problem for the regionalization because the regions need not to be disjoint.)
\end{itemize}

In a situation where the whole input space $\X$ is divided by a regionalization method into some regions $\X_1,\ldots,\X_B$ --- which need not to be disjoint --- we learn the SVMs separately, one SVM for each region. After that, we combine these local SVMs to a composed estimator or classifier, respectively. The influence of the local predictors may be  controlled pointwise by measurable weight functions $w_b : \X \to [0,1], b\in\btoB$, which have to fulfil the following two conditions for all $x\in\X$: \textbf{(W1)} $\sum_{b = 1}^B w_b(x)  = 1$ for all $x\in\X$, and \textbf{(W2)} $w_b(x) = 0$ for all $x\notin\Xb$  and for all $b\in\btoB.$

We follow the notation in \fullciteA{dumpert2018neuro} and define the composed predictors as follows:

\nomenclature[F]{$\fLPlambdacomp$}{Theoretical composed predictor in the overlapping case}
\nomenclature[F]{$\fLDnlambdacomp$}{Empirical composed predictor in the overlapping case}

\begin{align}\label{konstruktionvonfLPlambdacomp}
\fLPlambdacomp : \X\to\R, \ \ \ \ \  \fLPlambdacomp(x) :=\zusammentheoretischx,
\end{align}
\begin{align}
\fLDnlambdacomp : \X\to\R, \ \ \ \ \  \fLDnlambdacomp(x) := \zusammenempirischx,     \notag
\end{align}

where

\begin{itemize}

\item $P$ is the unknown distribution of $(X,Y)$ on $\X\times\Y$ and $\frakDn := n^{-1} \sum_{i=1}^n \delta_{(x_i, y_i)}$ is the empirical measure based on a sample or data set $\calDn := \left(  (x_1, y_1), \ldots, (x_n, y_n) \right)$ of $n$ i.i.d. realizations of $(X,Y)$.
\item $\Pb$ is the theoretical distribution on $\Xb\times \Y$, $\frakDnb$ its empirical analogon.  They are in fact probability distributions in all interesting situations, i.\,e. if $ P(\Xb\times \Y) > 0$ or $\frakDn(\Xb\times \Y) > 0$, respectively, because they are built from $P$ and $\frakDn$ as follows:

$$ \Pb := \begin{cases}
 \ \  {P(\Xb\times \Y)}^{-1} \ P_{|_{\Xb\times \Y}} \ \   &, \  \text{if} \ \  P(\Xb\times \Y) > 0         \\
  \ \ \ \ \ \ \ \ 0                                                                 &, \  \text{otherwise}
\end{cases}$$
and
$$\nomenclature[D]{$\frakDnb$}{Empirical distribution on $\Xb\times\Y$} \frakDnb := \begin{cases}
 \ \  {\frakDn(\Xb\times \Y)}^{-1}\  \frakDn_{|_{\Xb\times \Y}}  \ \   &, \  \text{if} \ \  \frakDn(\Xb\times \Y) > 0         \\
   \ \ \ \ \ \ \ \ 0                                                                                    &, \  \text{otherwise}
\end{cases}.$$

We  write $\frakDn(\Xb\times \Y) =|\calDnb| =:  n_b$. \nomenclature[C]{$n_b$}{Number of observations in $\Xb\times\Y$, $n_b = \vert \calDnb \vert$}
\item In an analogous way,  the regional marginal distribution of $X$ is defined by $\PbX := P^{\X}(\Xb)^{-1} P^{\X}_{|\Xb}$ if $P^{\X}(\Xb)>0$ and 0 otherwise.
\item $\lambda := (\lambda_1, \ldots, \lambda_B) \subset\  ]0, \infty[^{B},$ or, if we want to emphasize the number of data points, also $\lambda_n := \lambda_{(n_1,1)},\ldots, \lambda_{(n_B,B)}),\ n = \sum_{b=1}^B \ n_b,$ instead of a fixed $\lambda$.
\item  By $\lokaltheoretisch$ we denote the theoretical local SVM on $\Xb\times\Y$ with respect to $\Ls$ and $\Pb$, if $\Pb$ is a probability measure; if $\Pb$ is the null measure, $\lokaltheoretisch$ is an arbitrary measurable function. By $\lokalempirisch$ we denote the empirical local SVM learned on $\Xb\times\Y$ with respect to $\Ls$ and $\frakDnb$, if $\frakDnb$ is a probability measure; if $\frakDnb$ is the null measure, $\lokalempirisch$ is an arbitrary measurable function.

\end{itemize}

In the situation of a predictor composed of locally learned SVMs, this predictor is universal {(risk-)}  consistent, too. We recall the relevant theorem from \fullciteA{dumpert2018neuro}.

\begin{thm}\label{consistencytheorem}

Let $\X$ be a separable metric space. Let $L$ be a convex, Lipschitz continuous loss function (with Lipschitz constant $|L|_1\neq 0$) and $\Ls$ its shifted version.
For all $b\in\btoB$ let $k_b$ be a measurable and bounded kernel on $\X$ and let the corresponding RKHSs $H_b$ be separable. Let the regionalization method fulfil \textbf{(R1)}, \textbf{(R2)}, and \textbf{(R3)}.

Then for all distributions $P$ on $(\X\times\Y,\BorelXY)$ with $H_b$ dense in $L^1(\PbX), b\in\btoB,$ and every collection of sequences $\lambda_{(n_1,1)},\ldots,\lambda_{(n_B,B)}$ with $\lambdanb\to 0$  and $\lambda_{(n_b,b)}^{2}n_b \to\infty$ when $n_b\to\infty$, $b\in\btoB,$ it holds true that
$$\mathcal R_{\X,\Ls,P}(\fLDnlambdancomp) \nto \mathcal R_{\X,\Ls,P}^* \ \ \ \text{in probability with respect to $P$}.$$
\end{thm}


\section{Robustness in terms of the influence function}\label{results}

First, please note that there is already a robustness result in terms of the so-called maxbias shown in \fullciteA{dumpert2018neuro}. In this paper we use another notion of robustness, the so-called influence function according to \fullciteA{Hampel1968} considering a statistical operator $S$ which assigns to every distribution $ P$ on the Borel-$\sigma$-algebra $\mathfrak B_M$ of a suitable set $M$ an element of a Banach space, i.\,e. in the situation at hand the predictor $f_{\Ls, P,\lambda}$ (which is in the approach without regionalization even an element of a (reproducing kernel) Hilbert space).

\begin{dfi}\label{defIF}
The influence function of $S$ at a point $z$ for a distribution $P$ is (if it exists) 
$$\IF(\delta_z; S, P) :=\epslim\  \frac {S\left((1-\varepsilon) P + \varepsilon\delta_z\right) - S(P)}{\varepsilon}$$
where $\delta_z$ is the Dirac distribution at the point $z$.
\end{dfi}

The influence function can be interpreted in the way that it measures the impact of an infinitesimal small amount of contamination of the original distribution $P$ in direction of a Dirac distribution in the point $z$ on the quantity of interest $S(P)$. If the influence function exists and if it is continuous and linear, then it is a Gâteaux derivative of the operator $S: \mathcal M_1(\X\times\Y,\BorelXY)\to H,  P\mapsto f_{\Ls, P,\lambda}$ in the direction of the mixture distribution $(1-\varepsilon) P + \varepsilon\delta_z$. From this point of view, we are interested in conditions where our statistical method has a bounded influence function: the lower the bound, the more robust the method. Note that in this context IF itself is a function mapping a Dirac measure $\delta$ on $(\X\times\Y, \BorelXY)$ to a predictor in a RKHS, i.\,e. $\IF(\delta; S, P)(\cdot)\in H$. Therefore we can evaluate $\IF(\delta; S, P)(\cdot)$ at a point $x\in\X$ to receive a real value ($\IF(\delta; S, P)(x)\in\R$ for all $x\in\X$) due to Proposition~\ref{stetigerkern} if we use a continuous and bounded kernel.

\begin{prp}\label{globaleexistenz}
As shown in \fullciteA{christmann2009consistency} the influence function (in the unregionalized situation) exists and is bounded if $\X$ is a complete, separable metric space, $H$ is an RKHS of a bounded and continuous kernel $k$, $L$ is a convex and Lipschitz continuous loss function with continuous partial (Fréchet-)derivatives (with respect to the last argument) $L^\prime(y,\cdot)$ and $L^{\prime\prime}(y, \cdot)$ with $\sup_{y\in\Y} \|L^\prime(y, \cdot)\|_\infty \in\ ]0,\infty[$ and $\sup_{y\in\Y} \|L^{\prime\prime}(y, \cdot)\|_\infty < \infty$. The upper bound of the influence function in $H$-norm is given by $2\;\lambda^{-1}\|k\|_\infty|L|_1$. In sup-norm the upper bound is then $2\;\lambda^{-1}\|k\|_\infty^2 |L|_1$  according to Proposition~\ref{dieungleichung}.
\end{prp}

As it is already the case in the proof of universal consistency in \fullciteA{dumpert2018neuro}, all assumptions can be verified without knowing anything about the underlying distribution $P$. As an example one might mention a standard scenario: $\X = \R^d$ for a $d\in\N$, Gaussian-RBF-kernel $k(x,\tilde x) = \exp\left(-\gamma^{-2}\|x-\tilde x\|^2_2\right), \ x,\tilde x \in \X,$ for a $\gamma>0$ and the logistic loss function for regression $L(y,t):=-\ln\left({4\exp(y-t)}{(1+\exp(y-t))^{-2}}\right)$ or for classification $L(y,t) := \ln(1+\exp(-yt))$, respectively. Note that these loss functions fulfil the required properties but lead --- unfortunately --- only to convex optimization problems (instead of quadratic problems with box constraints which result by using non-smooth loss functions like the hinge loss for classification or the $\varepsilon$-insensitive loss for regression). Nevertheless there are extensions of the proofs on robustness properties also for these non-smooth loss functions, see \fullciteA{bouligand2008}, \fullciteA{christmann2009consistency}, and \fullciteA{van2010review}, but we would not prove these extensions for the localized situation within this paper. In the global, i.\,e. not regionalized, situation, we can rewrite the influence function as follows:
$$\IF(\delta_z; S, P) =\epslim\  \frac {f_{\Ls, (1-\varepsilon) P + \varepsilon\delta_z,\lambda} - f_{\Ls,P,\lambda}}{\varepsilon}.$$
This is used to define an influence function of the composed predictor defined in (\ref{konstruktionvonfLPlambdacomp}). Recall that this composed predictor is --- in general --- not an element of a Hilbert space --- however, it is an element of ${L}^\infty(\PX)$ on $\X$ and by this an element of a Banach space if we use bounded kernels. Thus, Hampel's definition is suitable in the regionalized situation, too. Define $\IF^{comp}(\delta_z; S, P)$, i.\,e. the influence function of the composed predictor, straightforwardly as follows.
\begin{dfi}
The influence function of the composed predictor  as defined in (\ref{konstruktionvonfLPlambdacomp}) is (if it exists)
$$\IF^{comp}(\delta_z; S, P) :=\epslim\  \frac {f^{comp}_{\Ls, (1-\varepsilon) P + \varepsilon\delta_z,\lambda} - f^{comp}_{\Ls,P,\lambda}}{\varepsilon},$$
where $S:\mathcal M_1(\X\times\Y, \BorelXY) \to {L}^\infty(\PX)$, $S(P) = f^{comp}_{\Ls,P,\lambda}.$
\end{dfi}

Note that in this regionalized context $\IF^{comp}$ itself is a function mapping a Dirac measure $\delta$ on $(\X\times\Y, \BorelXY)$ to a predictor in the Banach space, i.\,e. $\IF^{comp}(\delta; S, P)(\cdot)\in {L}^\infty(\PX)$ (if we use a bounded kernel, see Proposition~\ref{dieungleichung}). It is possible to show that also a composed predictor as defined in (\ref{konstruktionvonfLPlambdacomp}) has a bounded influence function. To do this, we use the following notation:

$$ \Pbtildeepsilonz := \begin{cases}
 \ \  (1-\varepsilon) \Pb + \varepsilon \delta_z \ \   &, \  \text{if} \ \  z\in\Xb\times\Y         \\
  \ \ \ \ \ \ \ \  \Pb                                                                 &, \  \text{otherwise}
\end{cases}.$$
By this, $\Pbtildeepsilonz$ stands for the mixture distribution on $(\Xb\times\Y, \BorelXbY)$ if the SVM on $\Xb$ is affected by $\delta_z$. In all other situations,  $\Pbtildeepsilonz = \Pb.$ This notation is necessary to guarantee that a local SVM is always learned with respect to a probability measure. Note that the local influence function $\IF_b$ is 0 in all situations where $\Pbtildeepsilonz=\Pb$, $b\in\btoB$.

For pointwise defined functions $g:U\to\R$ with $U\supset \Xb$ we define $\|g\|_{\Xb\text{-}\infty} :=  \sup_{\substack{x\in\Xb}} g(x)$, $b\in\btoB$; if $g$ is not pointwise defined, $\|g\|_{\Xb\text{-}\infty} :=  \inf\left\{K\ge 0 \mid \ |g|\le K\  \PbX \text{-a.\,s.}\right\}$, $b\in\btoB$.

\begin{thm}[Existence]\label{existenceofcompif}
Let for all $b\in\btoB$ $\Xb$ be a complete, separable metric space, $H_b$ an RKHS of a bounded and continuous kernel $k_b$, and let $L$ be a convex and Lipschitz continuous loss function with continuous partial (Fréchet-)derivatives (with respect to the last argument) $L^\prime(y,\cdot)$ and $L^{\prime\prime}(y, \cdot)$ with $\sup_{y\in\Y} \|L^\prime(y, \cdot)\|_{\Xb\text{-}\infty} \in\ ]0,\infty[$ and $\sup_{y\in\Y} \|L^{\prime\prime}(y, \cdot)\|_{\Xb\text{-}\infty} < \infty$, $b\in\btoB$. Then, $\IF^{comp}(\delta_z;S,P)$ exists and is bounded.
\end{thm}

Note that fulfilling assumptions \textbf{(R1)} to \textbf{(R4)} is sufficient to produce such regions $\Xb$ out of an input space $\X$. Also note that continuous kernels are of course measurable, and that their corresponding RKHSs are separable, see \fullciteA[Lemma~4.33]{steinwart2008support}. Note that $\sup_{y\in\Y} \|L^\prime(y, \cdot)\|_{\Xb\text{-}\infty} \in\ ]0,\infty[$ already implies the Lipschitz continuity of $L$ with $|L|_1 \neq 0$. This is useful for a fair comparison of the assumptions of the different theorems on consistency (Theorem~\ref{consistencytheorem}) and robustness (Theorems~\ref{existenceofcompif} and~\ref{robustnesstheorem}).

\begin{Proof}[Theorem \ref{existenceofcompif}]
To show the result, we decompose the predictor.
\begin{align}
\IF^{comp}(\delta_z; S, P) & =\epslim\  \frac {f^{comp}_{\Ls, (1-\varepsilon) P + \varepsilon\delta_z,\lambda} - f^{comp}_{\Ls,P,\lambda}}{\varepsilon}\\ \notag
& = \epslim\  \frac {\bsumme w_b    f_{b, \Ls, \Pbtildeepsilonz ,\lambdab} - \bsumme w_b f_{b, \Ls,\Pb,\lambdab}}{\varepsilon} \\ \notag
& = \epslim\ \  \bsumme \ w_b \ \frac {    f_{b, \Ls, \Pbtildeepsilonz,\lambdab} -  f_{b, \Ls,\Pb,\lambdab}}{\varepsilon} \\ \notag
& = \bsumme \ w_b \ \ \epslim\ \ \frac {    f_{b, \Ls, \Pbtildeepsilonz,\lambdab} -  f_{b, \Ls,\Pb,\lambdab}}{\varepsilon} \\ \notag
& = \bsumme \ w_b \ \IF_b(\delta_z; S_b, \Pb),\notag
\end{align}
where $S_b$ is the local statistical operator on $\Xb\times\Y$, i.\,e. $S_b : \mathcal M_1(\Xb\times\Y,\BorelXbY) \to H_b$, $S_b(\Pb) =  f_{b, \Ls,\Pb,\lambdab}$. According to Proposition \ref{globaleexistenz} the influence function of every local SVM exists and is bounded. Thus, the above sum exists and is bounded, too.
\end{Proof}

The upper bounds of the local influence functions, see \fullciteA{christmann2009consistency}, can be used to give an upper bound of the influence function of the global predictor. Every local influence function $\IF_b$, $b\in\btoB$, is bounded by $\lambdab^{-1}\|k_b\|_{\Xb\text{-}\infty} |L|_1 \|P - \delta_z\|_{(\Xb\times\Y)\text{-TV}}$ in $H_b$-norm,
where $\|k_b\|_{\Xb\text{-}\infty} := \sup_{\substack{x\in\Xb}} \ \sqrt{k_b(x,x)}, b\in\btoB$, $H_b$ is the RKHS of $k_b$, and where   $\|\cdot\|_{(\Xb\times\Y)\text{-TV}}$ is the  total variation norm on the space of distributions on $(\Xb\times\Y,\BorelXbY)$, for details see, e.\,g., \fullciteA[p.~158]{DenkowskiMigorskiPapageorgiou2003}. Note that according to Proposition~\ref{dieungleichung} we get $\| \IF_b(\delta_z; S_b, \Pb) \|_{\Xb\text{-}\infty}\le \| \IF_b(\delta_z; S_b, \Pb) \|_{H_b} \| k_b\|_{\Xb\text{-}\infty}.$

\begin{thm}[Upper bound]\label{upperbound}
Under the assumptions of  Theorem \ref{existenceofcompif} it holds true that 
\begin{align}
\| \IF^{comp}(\delta_z; S, P)\|_{\infty}\le\   2\ |L|_1\ \bsumme \ \|w_b\|_{\Xb\text{-}\infty} \ \lambdab^{-1}\|k_b\|_{\Xb\text{-}\infty}^2,\label{IFabschaetzung}
\end{align}
where $\|\IF^{comp}(\delta_z; S, P)\|_\infty :=  \inf\left\{K\ge 0 \mid \ |\IF^{comp}(\delta_z; S, P)(\cdot)|\le K\  \PX \text{-a.\,s.}\right\}$, $b\in\btoB$.
\end{thm}

\begin{Proof}[Theorem \ref{upperbound}]
By Theorem \ref{existenceofcompif} we can straightforwardly prove an upper bound for the influence function of the global predictor using the triangle inequality and using that the weights $w_b$ vanish outside $\X_b$, $b\in\btoB$, 
\begin{align}\notag
\| \IF^{comp}(\delta_z; S, P)\|_{\infty} & = \left\| \bsumme \ w_b \ \IF_b(\delta_z; S_b, \Pb)\  \right\|_{\infty}\\ \notag
                                                     & \le\   \bsumme \left\|   w_b\  \IF_b(\delta_z; S_b, \Pb) \right\|_{\infty}\\ \notag
                                                     & \le\   \bsumme \left\|   w_b\  \IF_b(\delta_z; S_b, \Pb) \right\|_{\Xb\text{-}\infty}\\ \notag
                                                    & =\   \bsumme \ \|w_b\|_{\Xb\text{-}\infty} \ \left\| \IF_b(\delta_z; S_b, \Pb)  \right\|_{\Xb\text{-}\infty}\\ \notag
                                                     & \le\   \bsumme \ \|w_b\|_{\Xb\text{-}\infty} \ \| \IF_b(\delta_z; S_b, \Pb) \|_{H_b} \| k_b\|_{\Xb\text{-}\infty}\\ \notag
                                                     & \le\   \bsumme \ \|w_b\|_{\Xb\text{-}\infty} \ \lambdab^{-1}\|k_b\|_{\Xb\text{-}\infty}^2 |L|_1 \|P - \delta_z\|_{(\Xb\times\Y)\text{-TV}}\\ \notag
                                                     & \le\   2\ \bsumme \ \|w_b\|_{\Xb\text{-}\infty} \ \lambdab^{-1}\|k_b\|_{\Xb\text{-}\infty}^2 |L|_1. \notag
\end{align}
The last inequality is true due to a general (and very rough) upper bound on the total variation norm. The inequality before follows from \fullciteA[Theorem~12]{christmann2009consistency} using the representer theorem for support vector machines with (convex and Lipschitz continuous) shifted loss functions \fullcite[Theorem~7]{christmann2009consistency}.
\end{Proof}

Note that there is a trade-off between two important properties of statistical methods in general and SVMs in particular: One of the assumptions for consistency of the composed global predictor in Theorem \ref{consistencytheorem} is that $\lambdanb\to 0$ for all $b\in\btoB$. Having a look at Inequality (\ref{IFabschaetzung}) we see that the smaller $\lambdab$ the higher is the upper bound of the influence function. This means that there is a trade-off between consistency and robustness of predictors based on locally learned SVMs. (This trade-off exists for SVMs in general --- not only in the regionalization approach.) The same problem has already arisen for the upper bound of the maxbias in \fullciteA{dumpert2018neuro} and is well-known for ill-posed problems in general and also for other notions of robustness, see e.\,g. \fullciteA{robustnessvsconsistency2013}.

Following \fullciteA{robustnessconvexrisks2004} it is possible to show properties of the influence function not only for a Dirac measure $\delta_z$ but also for an arbitrary distribution $Q$ on $(\X\times\Y, \BorelXY)$. In the situation of a locally learned predictor, we can prove this, too. In analogy to $P$ and $\Pb$ we define
$$ Q_b := \begin{cases}
 \ \    {Q(\Xb\times \Y)}^{-1} \ Q_{|_{\Xb\times \Y}} \ \   &, \  \text{if} \ \  Q(\Xb\times \Y) > 0         \\
  \ \ \ \ \ \ \ \ 0                                                                 &, \  \text{otherwise}
\end{cases},$$
i.\,e. $Q_b$ is a probability measure on regions $(\Xb\times\Y,\BorelXbY)$ if the support of $Q$ has a part in $\Xb\times\Y$ and the null measure otherwise. Using this, we can define 
$$ \PbtildeepsilonQb := \begin{cases}
 \ \  (1-\varepsilon) \Pb + \varepsilon Q_b \ \   &, \  \text{if} \ \  Q_b\neq 0         \\
  \ \ \ \ \ \ \ \  \Pb                                                                 &, \  \text{otherwise}
\end{cases}.$$
Note again that on all regions $\Xb\times\Y$ where $\PbtildeepsilonQb = \Pb$ the local influence function $\IF_b$ is zero.

\begin{cor}\label{IFmitQexistence}
Under the assumptions of  Theorem \ref{existenceofcompif} $\IF^{comp}(Q;S,P)$ exists and is bounded with upper bound $2\ |L|_1\ \sum_{b=1}^B \ \|w_b\|_{\Xb\text{-}\infty} \ \lambdab^{-1}\|k_b\|_{\Xb\text{-}\infty}^2$ uniformly for all distributions $P$ and $Q$ on $(\X\times\Y,\BorelXY)$.
\end{cor}
\begin{Proof}[Corollary \ref{IFmitQexistence}]
The proof can be done analogously to the proofs of Theorems~\ref{existenceofcompif} and~\ref{upperbound} in consideration that the local influence functions exist and are bounded.
\end{Proof}

To compare this notion of robustness to another one, the so-called maxbias, we recall the corresponding theorem from \fullciteA{dumpert2018neuro}.

\begin{thm}\label{robustnesstheorem}
Let $\X$ be a separable metric space. Let $L$ be a convex, Lipschitz-continuous (with Lipschitz-constant $|L|_1\neq 0$) loss function and $\Ls$ its shifted version.
For all $b\in\btoB$ let $k_b$ be a measurable and bounded kernel and let the corresponding RKHSs $H_b$ be separable. Let the regionalization method fulfil \textbf{(R1)}, \textbf{(R2)}, and \textbf{(R3)}. Define --- for $\varepsilon_b\in [0,\frac 1 2[$ --- $N_{b,\varepsilon_b}(\Pb) := \left\{(1-\varepsilon_b)\Pb + \varepsilon_b Q_b\ \left|\ Q_b\in\mathcal M_1(\Xb\times\Y,\BorelXbY) \right.\right\}$, $b\in\btoB$, and $$N_\varepsilon(P) := \left\{Q \in \mathcal M_1(\X\times\Y,\BorelXY) \ \left|\ Q_b\in N_{b,\varepsilon_b}\ \text{ for all } b\in\btoB \right.\right\}.$$ Then, for all distributions $P$ on $(\X\times\Y,\BorelXY)$ and all $\lambda := (\lambda_1,\ldots,\lambda_B)\in \ ]0,\infty[^B$, it holds that
\begin{align}\label{maxbiasabschaetzung}\underset{Q \in N_{\varepsilon}(P)}{\sup} \ \left\|\fLQlambdacomp - \fLPlambdacomp\right\|_{\infty}\le 2\ |L|_1 \ \bsumme\ \|w_b\|_{\Xb\text{-}\infty}\   \frac {\varepsilon_b}{\lambda_b}\  \|k_b\|^2_{\Xb\text{-}\infty}.\end{align}
\end{thm}

This bound is also a uniform bound in the sense that it is valid for all distributions $P$ and all weighting schemes fulfilling \textbf{(W1)} and \textbf{(W2)}, i.\,e. $\sum_{b = 1}^B w_b(x)  = 1$  for all $x\in\X$ and \ $w_b(x) = 0$ for all $x\notin\Xb$ and for all $b\in\btoB.$ In contrast to robustness in terms of the influence function, we do not have to fulfil the assumption that the regions $\Xb$, $b\in\btoB$, are complete or that the loss function is differentiable in order to prove the upper bound of the maxbias. On the other hand, the proof of the existence of the local influence functions \fullcite[Theorem~10]{christmann2009consistency} uses an implicit function theorem on Banach spaces and needs the completeness assumption for $\Xb$ for all $b\in\btoB$ and the continuity assumption for the kernels $k_b$ to show that the therein appearing inverse exists.

\begin{bsp}
Let $d\in\N$, $\X = \R^d$, $k_b$ be a Gaussian-RBF-kernel, i.\,e. $k_b(x,x^\prime) = \exp\left(-\gamma_b^{-2} \|x-x^\prime\|_2^2\right), \gamma_b>0$, for all $b\in\btoB$, and let $L$ be the logistic loss for classification or regression. Then Theorem~\ref{upperbound} and Corollary~\ref{IFmitQexistence} provide the uniform upper bound $2\, \sum_{b=1}^B \lambda_b^{-1}$ for the influence function of the composed predictor.
\end{bsp}

We can compare not only the two mentioned notions of quantitative robustness (maxbias and influence function) but also robustness in the regionalized vs. in the unregionalized (i.\,e. in the global) case. In the latter one, there is only one region ($B = B_n = 1$). Using this in (\ref{IFabschaetzung}) or (\ref{maxbiasabschaetzung}), respectively, and compare this to Proposition~\ref{globaleexistenz} or \fullciteA[Theorem~12]{christmann2009consistency}, respectively, we see that we do not lose robustness by using localized SVMs instead of one global one. (Note that $\|w_b\|_{\Xb\text{-}\infty}$ uses to be 1, $b\in\btoB$. Otherwise there would be a region with no points $(x,y)\in\X\times\Y$ on its own, i.\,e. a region that shares all of its points with at least one other region. This seems to be unrealistic as an outcome of a regionalization method.)

\section{Summary}\label{summary}

By proving and discussing quantitative robustness properties of locally learned predictors we have refined our analysis on local  learning. We showed that quantitative robustness properties  of kernel-based methods like support vector machines  are conserved in the local approach. We see that there is no disadvantage of learning separate predictors, one for each region, and combining them from this  point of view. All of the results have been shown for all distributions and only under assumptions which are verifiable by the user.

\section*{Support}
The work was partially supported by grant CH~291/2-1 of the Deutsche Forschungsgemeinschaft (DFG).

\bibliographystyle{newapa}
\renewcommand{\bibname}{References}
\renewcommand{\refname}{References}
\addcontentsline{toc}{chapter}{\bibname}
\bibliography{myBibliographie}

%
%
%

\end{document}